\documentclass[conference]{IEEEtran}
\usepackage{amsmath,bm}
\usepackage{array}
\usepackage{amssymb,amsfonts}
\usepackage{booktabs}
\usepackage[all,arc]{xy}
\usepackage{enumerate}
\usepackage{mathrsfs}
\usepackage{titletoc}
\usepackage[noend]{algpseudocode}
\usepackage{graphicx}
\usepackage[caption=false,font=footnotesize]{subfig}
\usepackage{fixltx2e}
\usepackage{balance}
\usepackage{algorithm}
\usepackage{subfig} 

\usepackage{cite}
\usepackage{array}
\usepackage{fixltx2e}
\usepackage{url}
\algdef{SE}{Begin}{End}{\textbf{begin}}{\textbf{end}}

\begin{document}
\title {Visualization of Multi-Objective Switched Reluctance Machine Optimization at Multiple Operating Conditions with t-SNE}

\author{
\IEEEauthorblockN{Shen Zhang\IEEEauthorrefmark{1}, Shibo Zhang\IEEEauthorrefmark{2}, Sufei Li\IEEEauthorrefmark{3}, Liang Du \IEEEauthorrefmark{4}, and Thomas G. Habetler\IEEEauthorrefmark{1}}

\IEEEauthorblockA{\IEEEauthorrefmark{1}\textit{School of Electrical and
Computer Engineering}, 
\textit{Georgia Institute of Technology}, 
Atlanta, GA 30332, USA\\
} 
\IEEEauthorblockA{\IEEEauthorrefmark{2}\textit{Computer Science}, 
\textit{Northwestern University}, 
Evanston, IL 60201, USA\\
}
\IEEEauthorblockA{\IEEEauthorrefmark{3}\textit{Ansys Inc.}, 
Canonsburg, PA 15317, USA\\
}
\IEEEauthorblockA{\IEEEauthorrefmark{4}\textit{Department of Electrical Engineering}, 
\textit{Temple University}, 
Philadelphia, PA 19122, USA \\
} 
}

\date{}
\maketitle
\thispagestyle{empty}
\pagestyle{empty}

\renewcommand{\figurename}{Fig.}


\maketitle

%
\begin{abstract}

The optimization of electric machines at multiple operating points is crucial for applications that require frequent changes on speeds and loads, such as the electric vehicles, to strive for the machine optimal performance across the entire driving cycle. However, the number of objectives that would need to be optimized would significantly increase with the number of operating points considered in the optimization, thus posting a potential problem in regards to the visualization techniques currently in use, such as in the scatter plots of Pareto fronts, the parallel coordinates, and in the principal component analysis (PCA), inhibiting their ability to provide machine designers with intuitive and informative visualizations of all of the design candidates and their ability to pick a few for further fine-tuning with performance verification. Therefore, this paper proposes the utilization of t-distributed stochastic neighbor embedding (t-SNE) to visualize all of the optimization objectives of various electric machines design candidates with various operating conditions, which constitute a high-dimensional set of data that would lie on several different, but related, low-dimensional manifolds. Finally, two case studies of switched reluctance machines (SRM) are presented to illustrate the superiority of then t-SNE when compared to traditional visualization techniques used in electric machine optimizations.

\end{abstract}
%
%
%
\IEEEpeerreviewmaketitle
\section{Introduction}
%
%

%
The process of electric machine design is a complex mixture of multi-physics field interactions and multi-objective optimizations \cite{YD_review}. In the recent years, there is also an increasing demand to optimize these machines at multiple operating points \cite{Multi_OP1, Multi_OP2} for applications that require frequent changes of speeds and loads, such as an electric vehicles with driving cycles, in which the objectives at different operating points may be in conflict with each other, and the overall dimension of objectives will increase substantially. 

Traditionally, the most commonly used methods for electric machine designs are evolutionary algorithms with a Pareto-based fitness assignment. Despite their success, the difficulty of solving multi-objective optimization problems increases with the number of objectives. In addition, presenting and visualizing the solution set of a many-objective problem (with four or more objectives) could end up becoming problematic \cite{vis_review1, vis_review2}, as most of the design candidates would become non-dominated and the Pareto-ranking will no longer work as a good discriminator. Moreover, even if sufficient solutions were generated via either a simple exhaustive search of design parameters or more intelligent search algorithms, it is difficult to present and visualize them in such a hyper-dimensional objective space, and therefore even harder for machine designers who are attempting to select the most appropriate candidates from the solution set for a targeted application.

A pertinent literature survey reveals that the data visualization in many-objective electric machine design is still an under-explored domain. While the scatter plots and parallel coordinates can be logically straightforward, distinguishing between the the data points on these plots may become difficult; when dealing with elements of large dimensions in a solution set. In addition, both of the above algorithms do not offer clustering or dimension reduction of the dataset, making it even harder to be implemented on electric machines optimized at multiple operating points. One of the traceable work employing a non-classic visualization tool in the field of electric machine designs uses an Aggregate Tree (AT) \cite{AT}, in conjunction with the parallel coordinates to assist in the progressive preference articulation, aiding the decision making process of an interior permanent magnet synchronous motor design. The results show that the ``AT is able to provide insight into the electrical machine design problem (in accordance with the common knowledge of physics) as well as guidance in the reduction of objectives". In addition, the use of self-organizing maps (SOM) is presented in \cite{Shen_SOM} to effectively cluster and visualize switched reluctance machines with four objectives. However, the dimension of objectives for machine design candidates that can be properly addressed by these techniques are still limited and their visualization performance will be compromised with hundreds and thousands of design candidates at multiple operating conditions, which is a common issue that machine designers may expect in the industry.

In this context, this paper proposes a methodology employing t-distributed stochastic neighbor embedding (t-SNE) to assist the visualization and data mining of the electric machine design solution set, which can be used by machine designers and engineers to better understand the relationships between the different objectives, and to then facilitate them to make the most appropriate pick in a more effective way. In this paper, the strength of t-SNE when compared to traditional visualization techniques, such as PCA and Isomap, is demonstrated using two case studies on a high-speed switched reluctance machine (SRM).
%

%
%
%
\section{Establishing the t-SNE Framework Visualizing Electric Machine Candidates}
\begin{figure}
\centering
\subfloat[]{\includegraphics[width=2.8in]{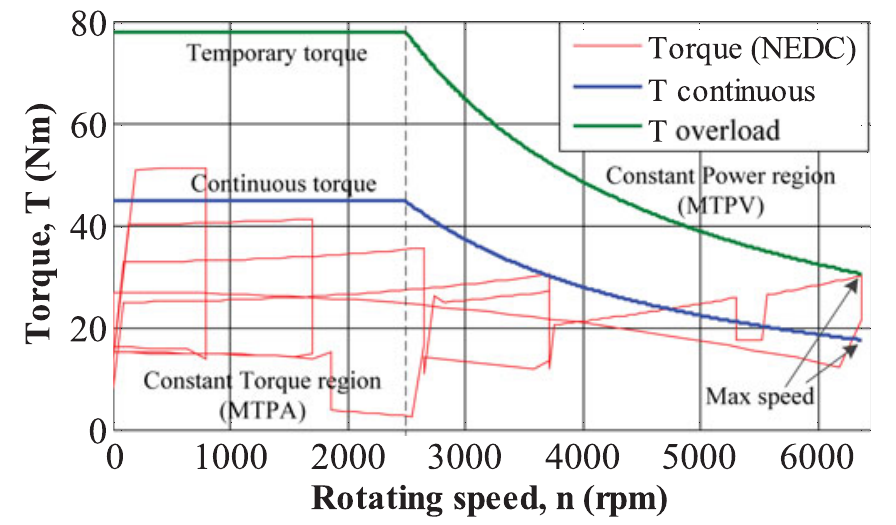}} 
\hspace{0.7in}
\subfloat[]{\includegraphics[width=3.0in]{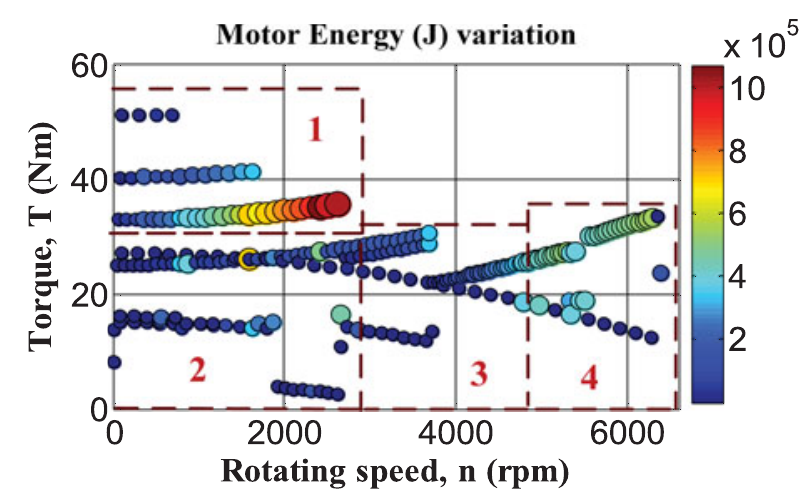}} 
\caption{(a) Illustration of the New European Drive Cycle (NEDC) for an electric vehicle (red line) and its torque-speed curve; and (b) motor energy distribution and clustering over the NEDC with torque and rotating speed \cite{Multi_OP2}.}
\label{Mul-OP} 
\end{figure} 
\subsection{Electric Machine Optimization: Single Operating Point VS Multiple Operating Points}
After performing an optimization process on electric machines for targeted applications, the design candidates form a collection of $N$ high-dimensional objects $(\mathbf { x } _ { 1 } , \mathbf { x } _ { 2 }, \dots , \mathbf { x } _ { N })\in \mathbb{R}^D$, where $D$ is the number of objectives to be optimized. For commonly applied stochastic optimization algorithms, such as the particle swarm optimization (PSO), genetic algorithms (GA), and differential evolution (DE) \cite{YD_review}, $N$ is determined by the initial population size and the number of iterations, which can vary from a few hundred to hundreds of thousands. 

When an electric machine is only optimized for a single operating point, the number of dimensions, $D$, is usually not large (usually around 3 to 8 based on the references in \cite{YD_review}), as the objectives associated with electric machines are typically the average torque, torque ripple, efficiency, torque density, machine weight, volume, quantified measures of the manufacturing complexity such as the stator tooth-slot shape \cite{Minos1}, as well as permanent magnet (PM) relevant metrics for PM machines such as the PM cost, volume, and its demagnetization vulnerability with various faults. With a relatively small $D$, the size of design candidates, $N$, can be also well regulated, since it is still easy to construct effective Pareto fronts when the number of objectives $D$ is small. In this scenario, classical visualization approaches such as the scatter plots of Pareto fronts, parallel coordinates and PCA should still be able to provide useful insights and intuitions to guide the next-stage fine-tuning and decision-making process.

However, for certain applications where the electric machine is experiencing frequent changes in speed and load, such as in electric vehicles that follow some volatile and unpredictable driving cycles, such as those illustrated in Fig. \ref{Mul-OP}(a), then this driving cycle will be visualized in different clusters, the centroids of which will serve as the representative operating points to optimize an electric machine, as can be seen in Fig. \ref{Mul-OP}(b), which has 4 clusters. This multi-operating point based optimization ensures an overall optimal performance across the entire driving cycle \cite{Multi_OP1, Multi_OP2}. If the number of operating point is $M$ (typically greater than 2), $D$ is redefined as the number of objectives for a single operating point, then the design candidates will form a new collection of $N$ high-dimensional objects $(\mathbf { x } _ { 1 } , \mathbf { x } _ { 2 }, \dots , \mathbf { x } _ { N })\in \mathbb{R}^{MD}$. In this case, the PCA is less likely to generate satisfactory visualization results, not to mention the scatter plots or the parallel coordinates, due to the inherent limitations of the algorithms themselves, as will be explained in the next subsection. In addition, t-SNE will be introduced to visualize these machine design candidates optimized for multiple operating points.

\begin{figure*}
\centering
\subfloat[]{\includegraphics[width=3.4in]{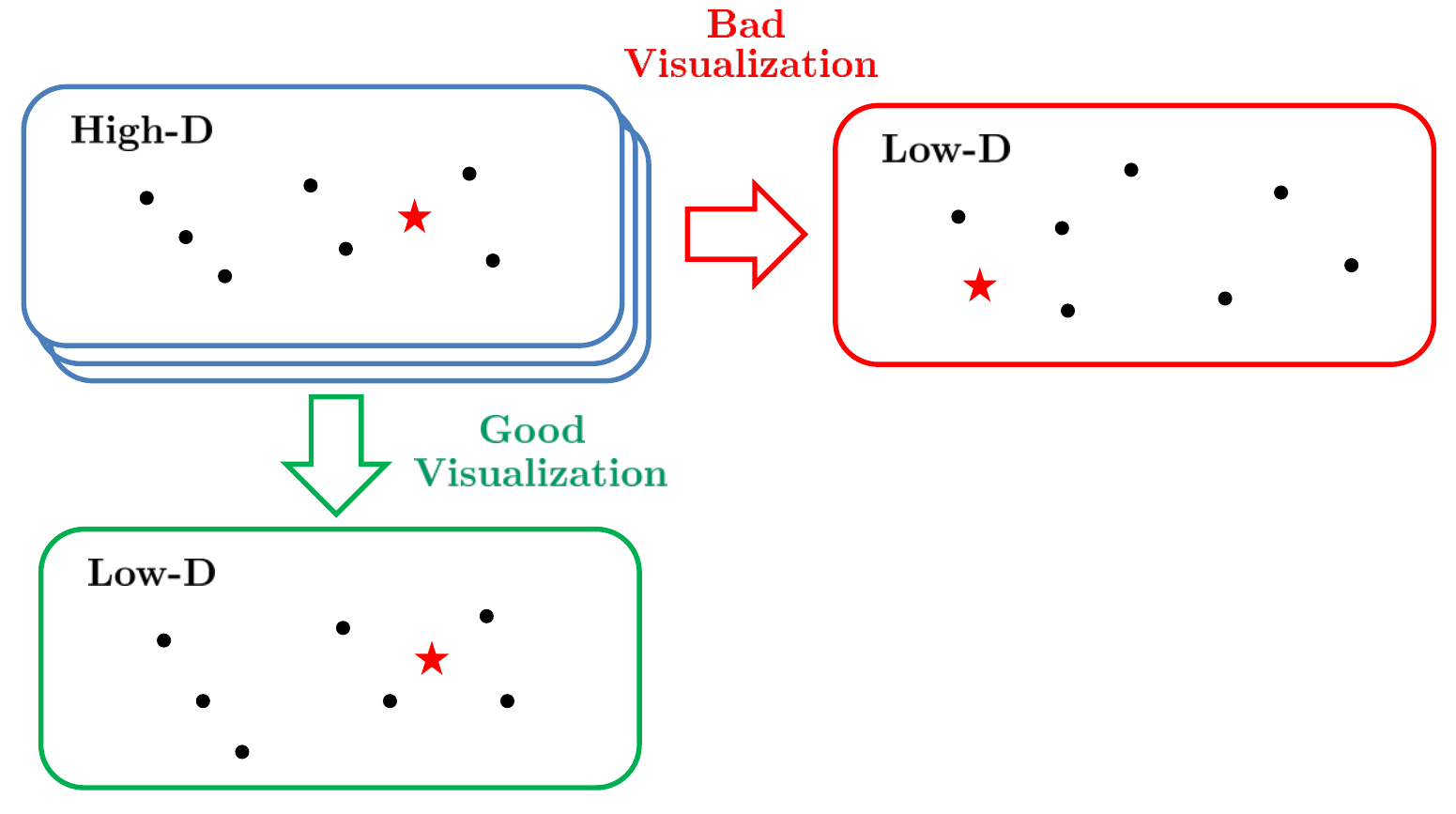}} 
\hspace{0.6in}
\subfloat[]{\includegraphics[width=2.0in]{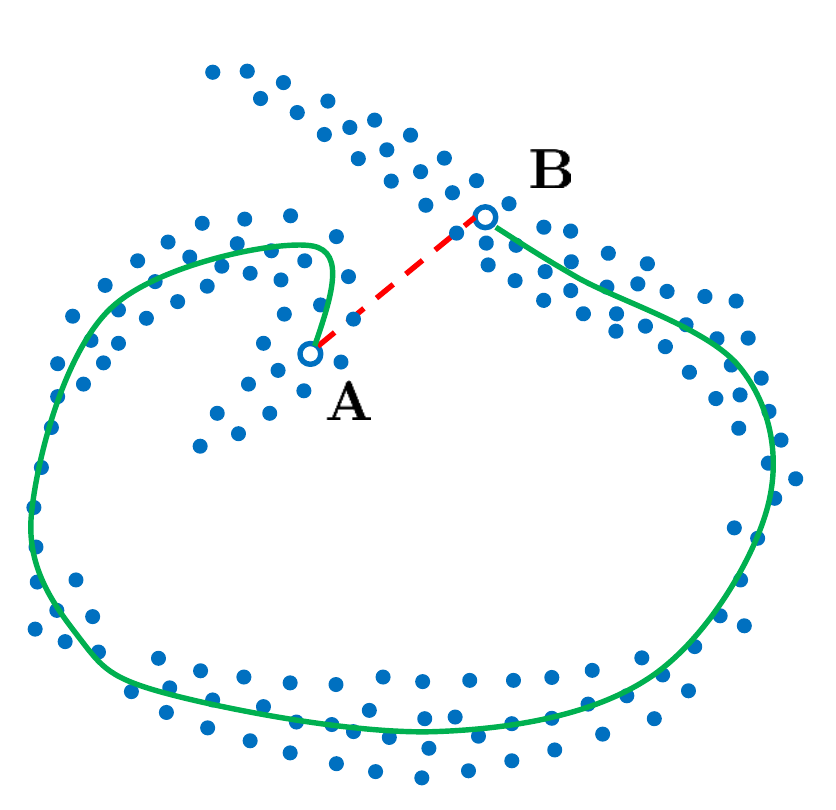}} 
\caption{(a) Illustration of projecting the data points from the high-dimensional space to the low-dimensional map; and (b) example of a nonlinear manifold in which the Euclidean distance fails to reflect the actual pairwise similarities between points.}
\label{Projection} 
\end{figure*} 
%
%

%
%
\subsection{Preserving Local Similarities in Visualizations}
For visualizing the $N$ high-dimensional electric machine design candidates $(\mathbf { x } _ { 1 } , \mathbf { x } _ { 2 }, \dots , \mathbf { x } _ { N })\in \mathbb{R}^{MD}$, it is desirable to obtain a good level of intuition for how these design candidates are arranged in the data space to facilitate the later decision making processes, for example, how many clusters they form, the local structure of the data manifold, etc. While the traditional visualization methods, such as parallel coordinates and scatter plots can indeed provide some simple and nice plots, they can only effectively visualize a few dimensions at once. 

Another popular method of visualizing data is to form a projection from the high-dimensional space to a low-dimensional map, where the distances between points reflect the similarities in the data. A good projection needs to properly preserve the point-wise distances in such a way that the low-dimensional map can accurately reflect the original high-dimensional space, as shown in Fig. \ref{Projection}(a). To do this, it is generally necessary to minimize some objective functions that measure the discrepancy between the similarities (distances) in the original high-dimensional data and in the low-dimensional map.

One technique of this distance-based visualization uses the principal component analysis (PCA), which attempts to find the first principal component by minimizing the \emph{linear} projection errors while  simultaneously maximizing the variance of the projected data. However, since PCA only examines the linear Euclidean distances between points, for some high-dimensional data that are more likely to form nonlinear manifolds, the Euclidean distance between points would not adequately reflect their similarity, as depicted in Fig. \ref{Projection}(b). The Euclidean distance (red dotted line) suggests that points A and B are similar, whereas they are actually very far apart when considering the entire manifold (green solid line). In addition, PCA tends to preserve the large pairwise distances over the small ones, since the low-dimensional subspace is found with maximal variance, indicating this subspace will tend to be aligned close to points lying far away from the center. 

Despite its simplicity and popularity, PCA in fact does not work well for visualization, since it only preserves large pairwise distances that are not reliable. Rather, the very small pairwise distances between points and their nearest neighbors can accurately preserve the local similarities, even with very curved data manifolds, as can be seen in Fig. \ref{Projection}(b). There has thus been an evolution of visualization techniques during the last 20 years, as evidenced by improved algorithms such as the Isomap \cite{Isomap}, locally linear embedding \cite{LLE}, stochastic neighbor embedding (SNE) \cite{SNE}, and t-SNE \cite{t-SNE}. 
\begin{figure*}
\centering
{\includegraphics[width=5.4in]{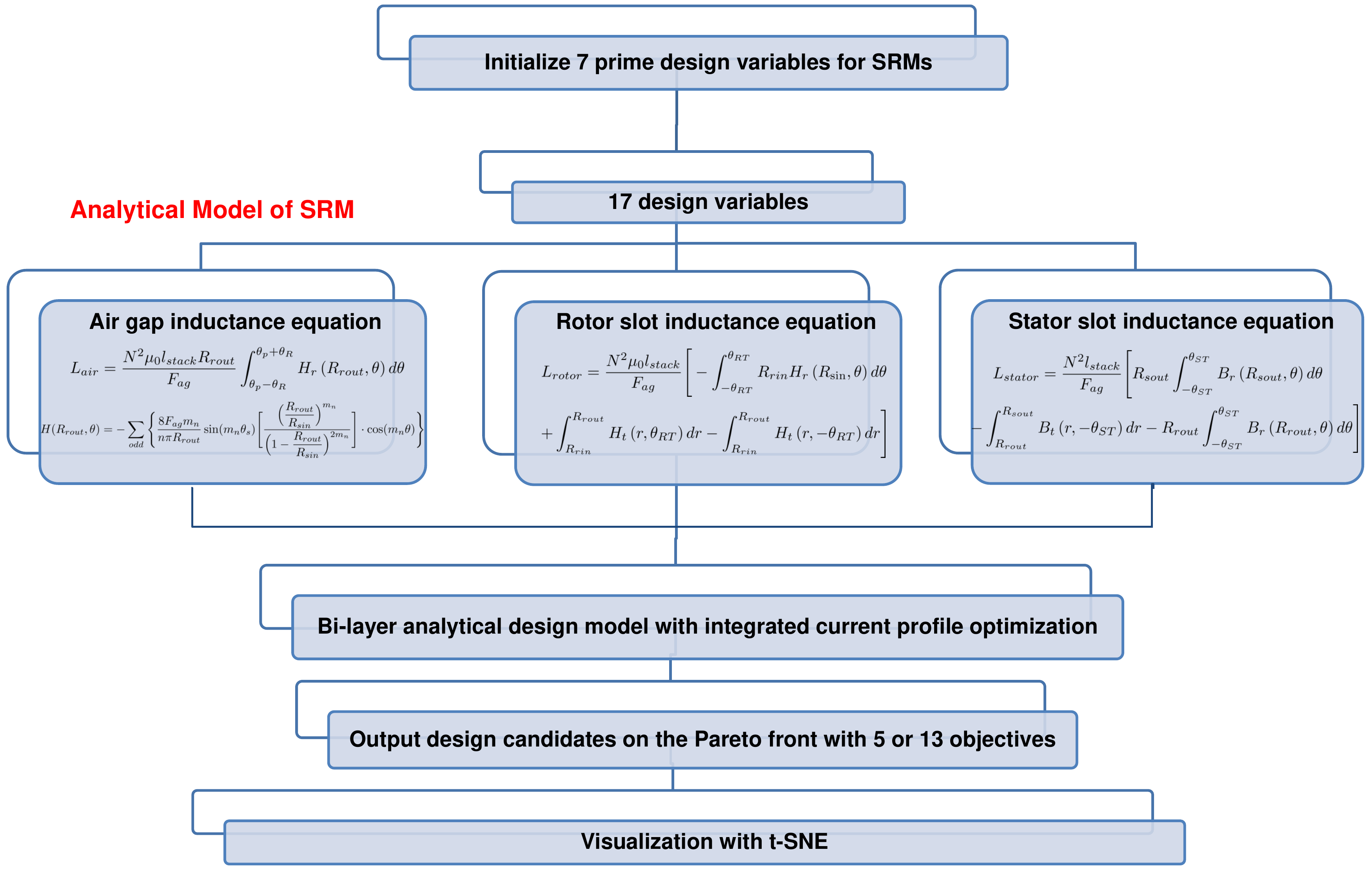}} 
\caption{The proposed analytical design, optimization and visualization process of switched reluctance machines.}
\label{chart} 
\end{figure*} 

\subsection{The Formulation of t-SNE}
The t-SNE algorithm was proposed in 2008 \cite{t-SNE} and has become one of the most popular high-dimensional data visualization techniques over the past decade. The algorithm assigns each data point a location in a two or three-dimensional map. This technique is a variation of Stochastic Neighbor Embedding (SNE), and is much easier to optimize, and produces significantly better visualizations by reducing the tendency to crowd points together in the center of the map. The t-distributed stochastic neighbor embedding is better than existing techniques at creating a single map that reveals structure at many different scales. This is particularly important for high-dimensional data that lie on several different, but related, low-dimensional manifolds, such as images of objects from multiple classes seen from multiple viewpoints. For visualizing the structure of very large data sets, t-SNE can use random walks on neighborhood graphs to allow the implicit structure of all of the data to influence the way in which a subset of the data is displayed. The performance of t-SNE can be seen on a wide variety of data sets and compared with many other non-parametric visualization techniques, including Sammon mapping, Isomap, and Locally Linear Embedding. The visualizations produced by t-SNE are significantly better than those produced by the other techniques on nearly all of the data sets, including MNIST dataset, CIFAR-10 image dataset, and TIMIT speech dataset, and street view house numbers on Google map, etc.

The t-SNE algorithm utilizes a joint probability distribution to model the similarity in high-dimensional space,
\begin{equation} p _ { i j } = \frac { \exp \left( - \left\| \mathbf { x } _ { i } - \mathbf { x } _ { j } \right\| ^ { 2 }/ 2 \sigma ^ { 2 } \right) } { \sum _ { k } \sum _ { l \neq k } \exp \left( - \left\| \mathbf { x } _ { k } - \mathbf { x } _ { l } \right\| ^ { 2 }/ 2 \sigma ^ { 2 } \right) }. \end{equation}
where $\sigma$ is the variance parameter of Gaussian, which is obtained via a binary search that produces a probability distribution $P_i$ with a fixed perplexity $Perp$  specified by the user.

Moreover, to eliminate the ``crowding problem", t-SNE employs the ``student t-distribution" with one degree of freedom to model the similarity between data $y_i$ and $y_j$ in low-dimensional space as
\begin{equation} q _ { i j } = \frac { \left( 1 + \left\| \mathbf { y } _ { i } - \mathbf { y } _ { j } \right\| ^ { 2 } \right) ^ { - 1 } } { \sum _ { k } \sum _ { l \neq k } \left( 1 + \left\| \mathbf { y } _ { k } - \mathbf { y } _ { l } \right\| ^ { 2 } \right) ^ { - 1 } }. \end{equation}

t-SNE finds the optimal low-dimensional representations for matching $p_{ij}$ and $q_{ij}$ to the greatest extent. This is achieved by minimizing the following Kullback-Leibler divergence measuring the difference between two probability distributions,
\begin{equation} C(Y)=KL(P||Q) = \sum_{i} \sum _{j\neq i}p_{ij}\log \frac {p_{ij}}{q_{ij}}. \end{equation}

t-SNE calculates the optimal low-dimensional representation $Y$ by minimizing $C(Y)$ over all data points with a gradient descent method, and the gradient of which is

\begin{equation}
    \frac { \delta C } { \delta y _ { i } } = 4 \sum _ { j } \left( p _ { i j } - q _ { i j } \right) \left( y _ { i } - y _ { j } \right) \left( 1 + \left\| y _ { i } - y _ { j } \right\| ^ { 2 } \right) ^ { - 1 }.
\end{equation}
which can be interpreted as a simulation of an \emph{N-body system}. 



%
%
\subsection{The t-SNE Algorithm}
The t-SNE algorithm is defined as follows by its inventor van der Maaten as shown in \textbf{Algorithm 1} \cite{t-SNE}, which is much easier to optimize, and ultimately yields significantly more useful visualizations than those produced by the other techniques. Besides its ability to preserve small pairwise distances while also not collapsing all points onto a single point by introducing the t-distribution that has a long tail than those in standard Gaussian Process, t-SNE can use random walks on neighborhood graphs of very large data sets, and allow the implicit structure of all of the data to influence the way in which a subset of the data is displayed. 
\begin{algorithm}[!t]
\caption{t-Distributed Stochastic Neighbor Embedding (t-SNE)}\label{algorithm_1}
\begin{algorithmic}
\State \textbf{Require}: the input data set $\chi = \left\{ x _ { 1 } , x _ { 2 } , \dots , x _ { n } \right\}$, perplexity $Perp$, number of iterations $T$, \\learning  rate $\eta$, momentum $\alpha (t)$.
\Begin
\State compute high-dimensional pairwise distances $p_{ij}$ with equation (1)
\State sample initial solution $\mathcal { Y } ^ { ( 0 ) } = \left\{ y _ { 1 } , y _ { 2 } , \dots , y _ { n } \right\}$ from $\mathcal { N } \left( 0,10 ^ { - 4 } I \right)$
\For {$t=1,2...T$}
\State compute low-dimensional pairwise distances $q_{ij}$ with equation (2)
\State compute gradient $\frac { \delta C } { \delta y }$ with equation (4) 
\State set  $\mathcal { Y } ^ { ( t ) } = \mathcal { Y } ^ { ( t - 1 ) } + \eta \frac { \delta C } { \delta y } + \alpha ( t ) \left( \mathcal { Y } ^ { ( t - 1 ) } - \mathcal { Y } ^ { ( t - 2 ) } \right)$
\EndFor
\State \textbf{end}
\End
\State \textbf{Results}: low-dimensional data representation $\mathcal { Y } ^ { ( T ) } = \left\{ y _ { 1 } , y _ { 2 } , \ldots , y _ { n } \right\}$.
\end{algorithmic}
\end{algorithm}
%
%



%
%
%
%
\section{Visualizing Multi-Objective SRM Design Candidates with t-SNE}
\subsection{Many-Objective Design and Optimization of SRMs}
\begin{figure}
\centering
\subfloat[]{\includegraphics[width=1.8in]{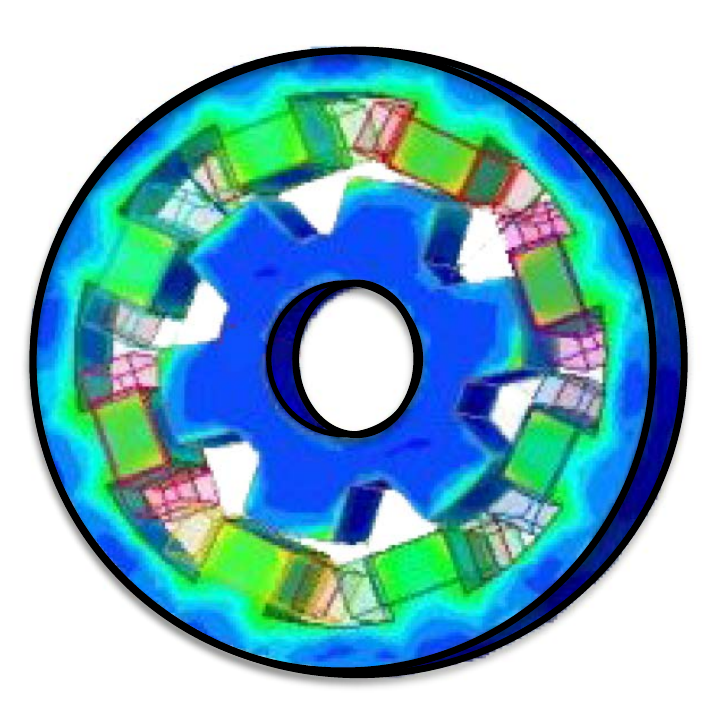}} 
\hspace{0.6in}
\subfloat[]{\includegraphics[width=2.1in]{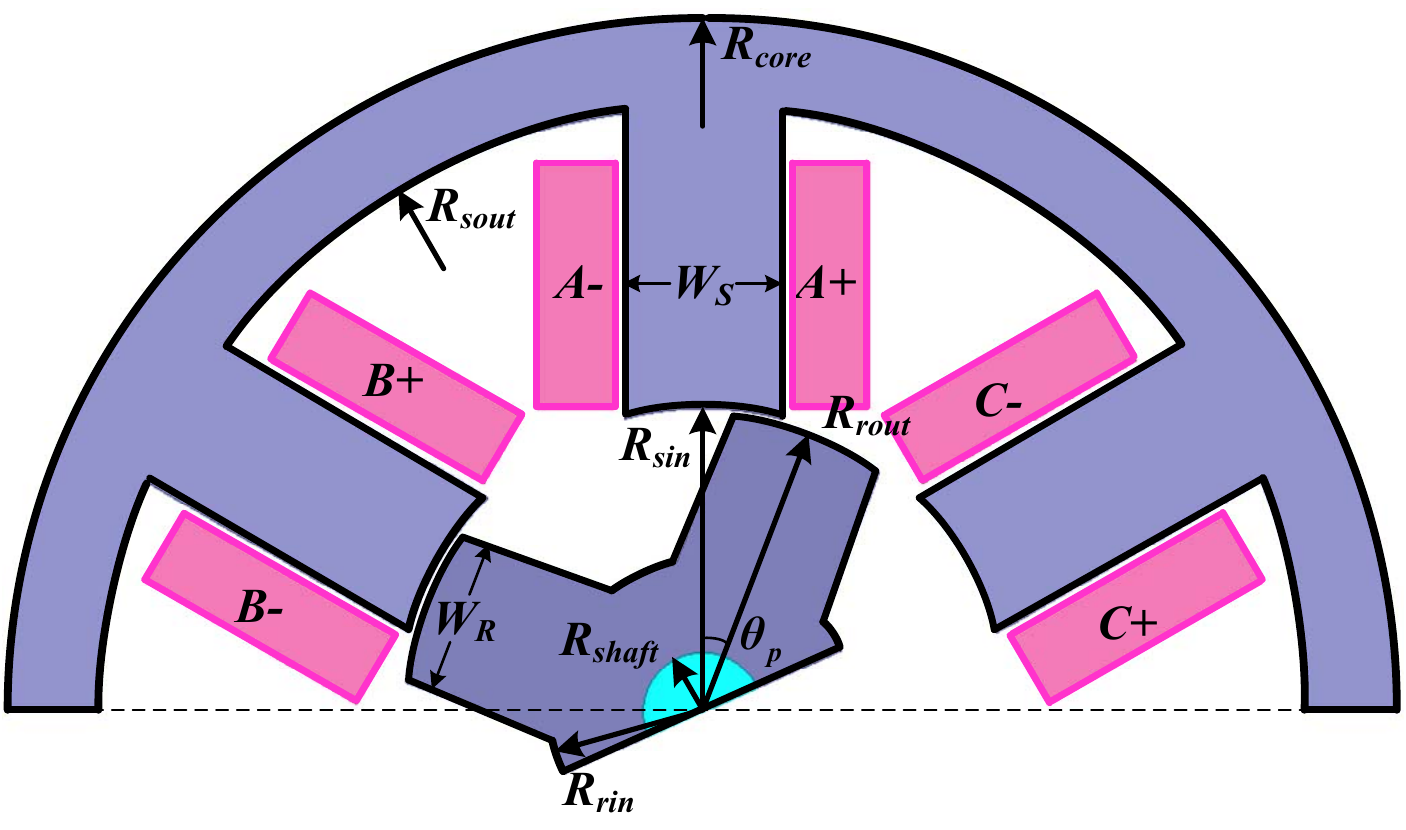}} 
\caption{Illustration of an SRM: (a) 3-D view and (b) cross-section view with geometric parameters.}
\label{SRM} 
\end{figure} 
\begin{figure*}
\centering
\subfloat[PCA]{\includegraphics[width=1.6in]{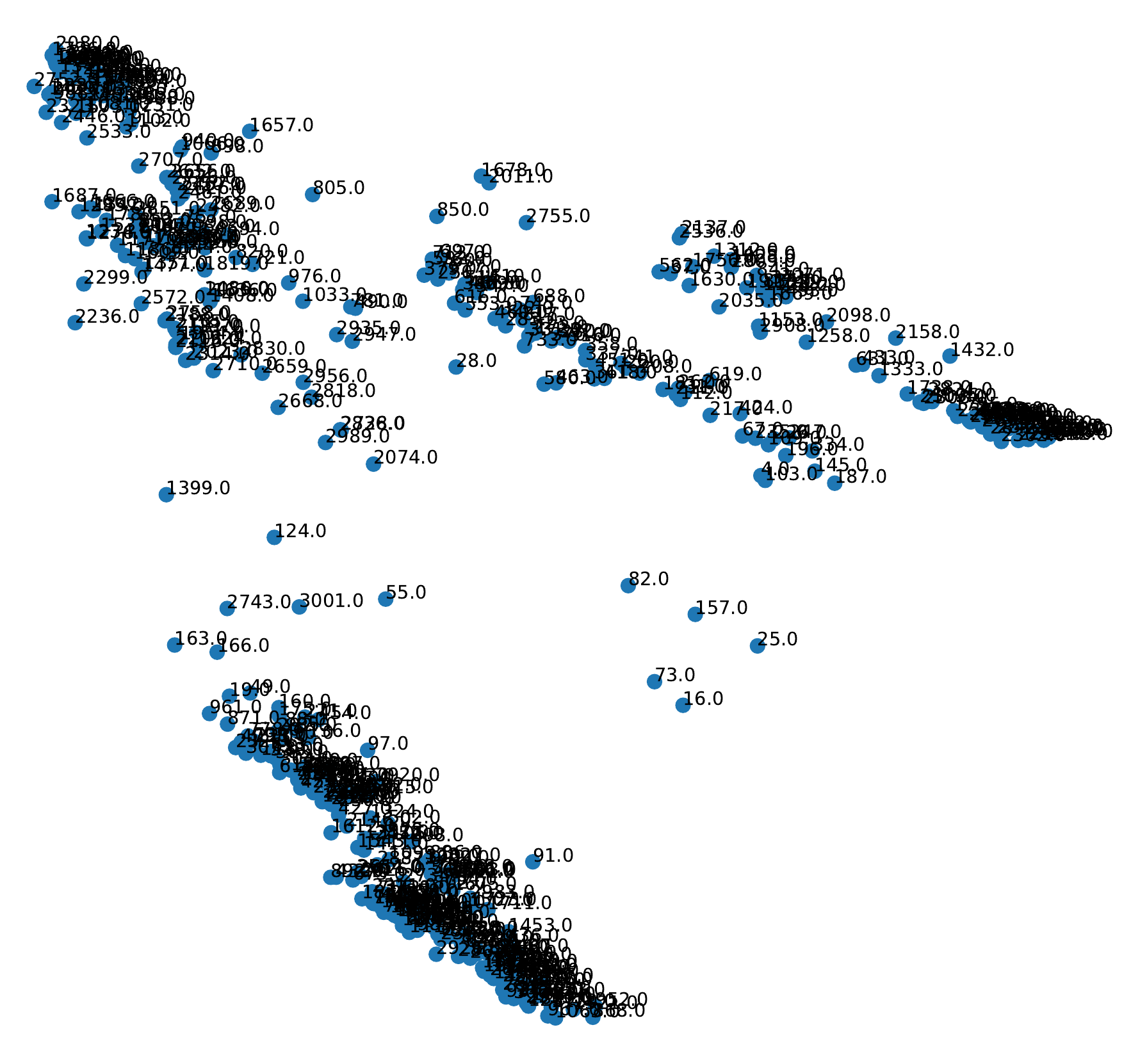}} 
\hspace{0.9in}
\subfloat[Isomap]{\includegraphics[width=0.85in]{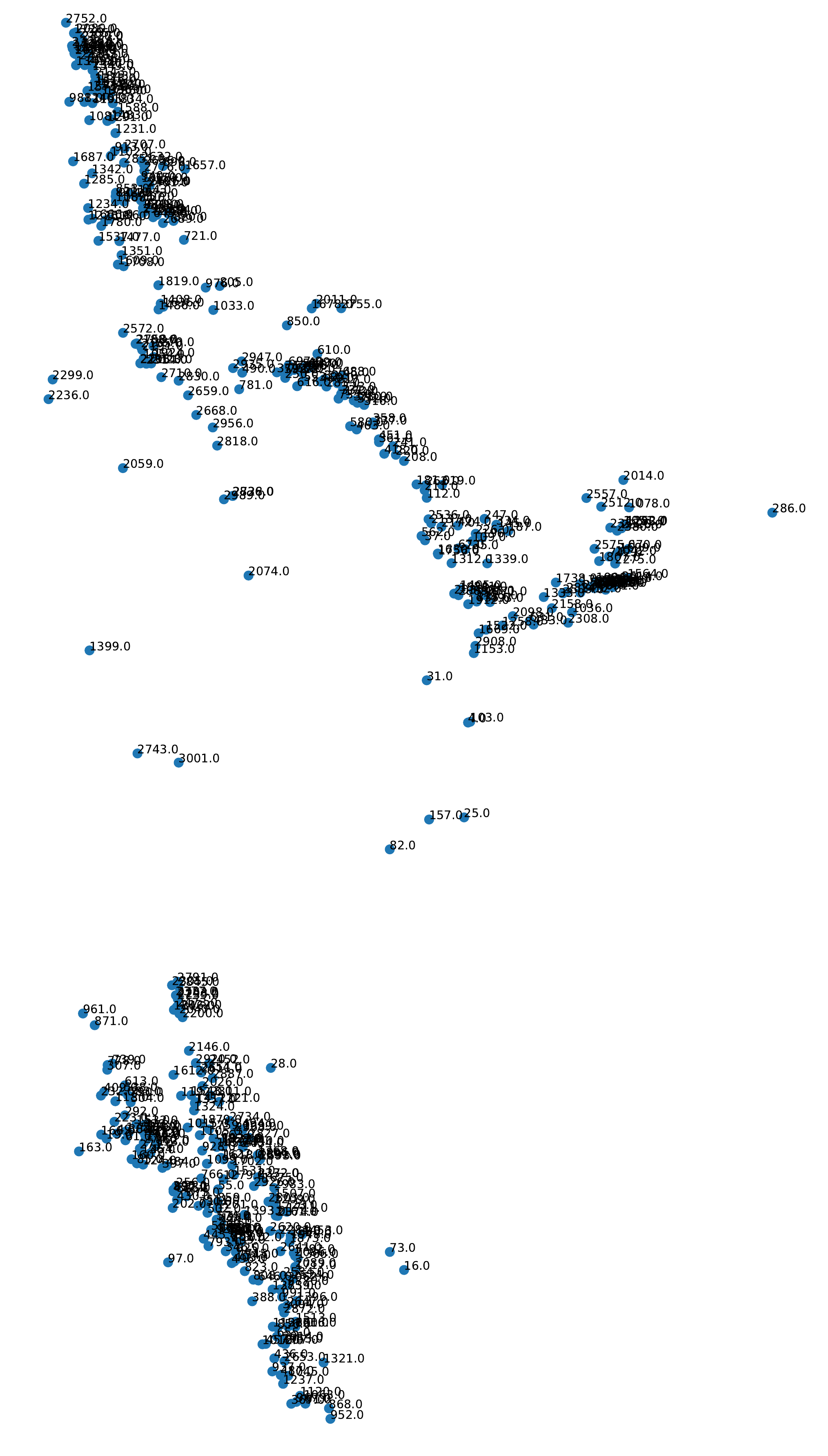}} 
\hspace{0.9in}
\subfloat[t-SNE]{\includegraphics[width=1.7in]{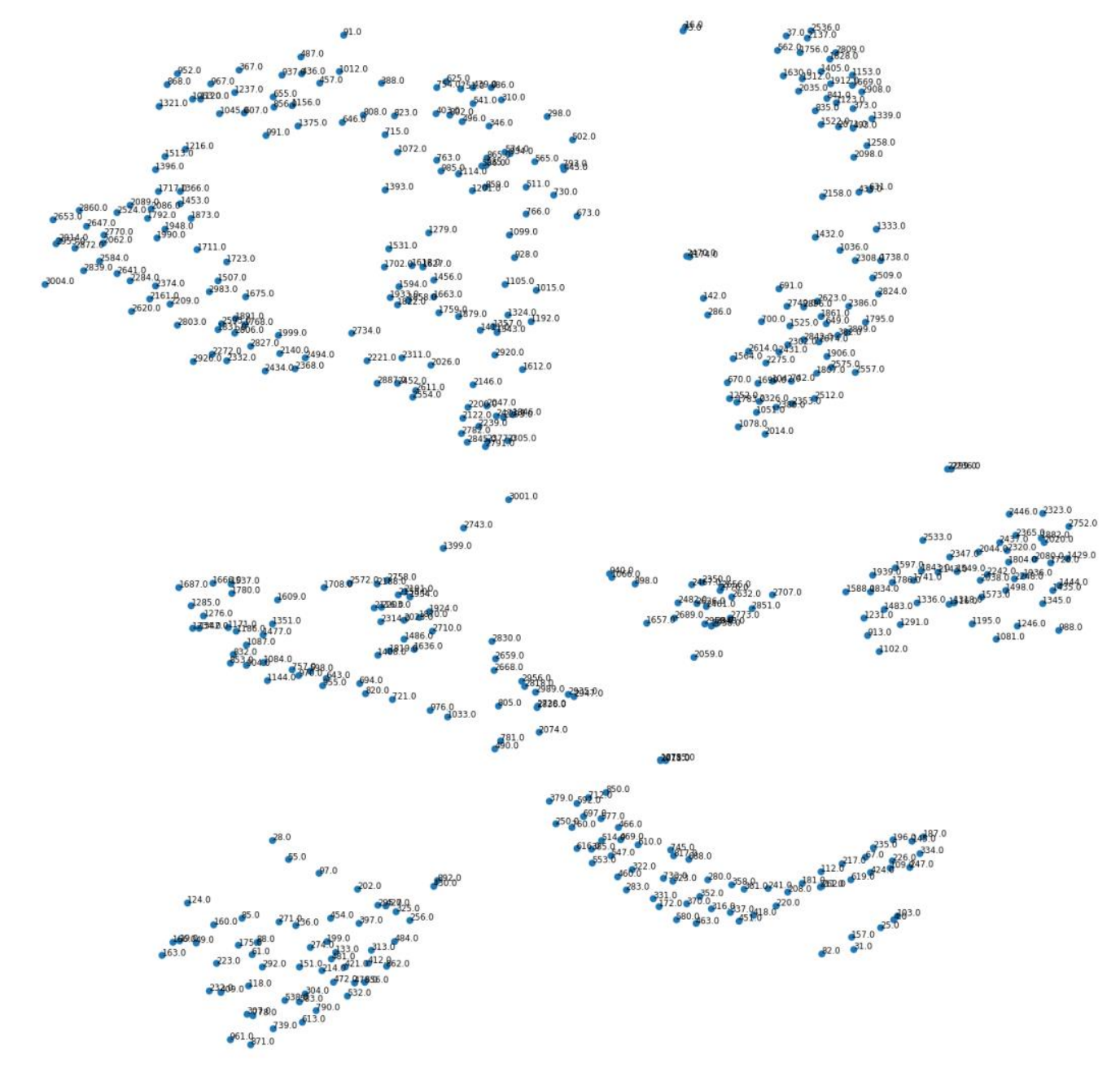}} 
\caption{Visualization of the 460 SRM design candidates optimized for a single operating point with 5 objectives: (a) visualization by PCA; (b) visualization by Isomap; and (c) visualization by t-SNE.}
\label{motor_6_dim} 
\end{figure*} 
\begin{figure*}
\centering
\subfloat[PCA]{\includegraphics[width=1.6in]{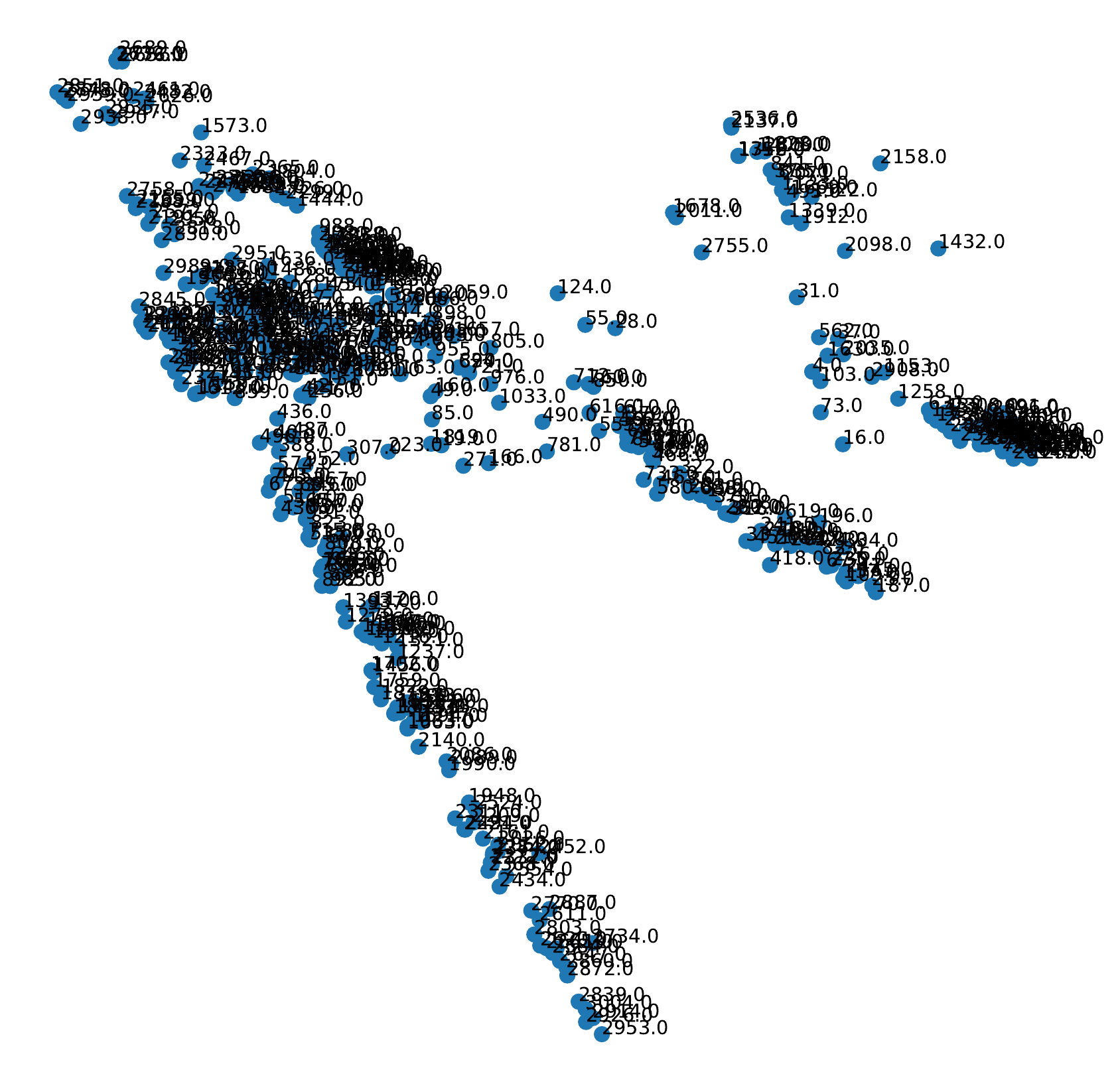}} 
\hspace{0.9in}
\subfloat[Isomap]{\includegraphics[width=0.75in]{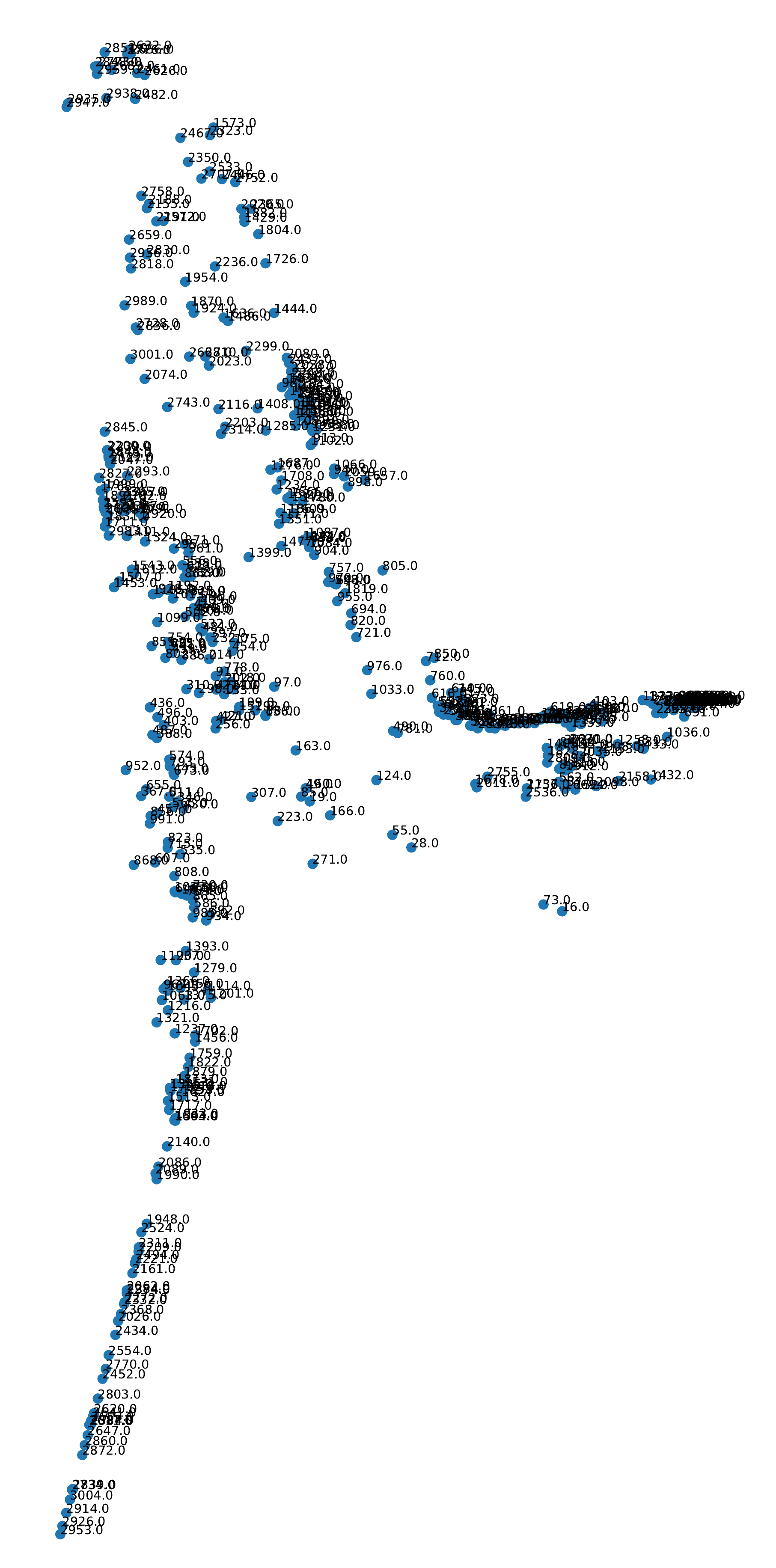}} 
\hspace{0.9in}
\subfloat[t-SNE]{\includegraphics[width=1.7in]{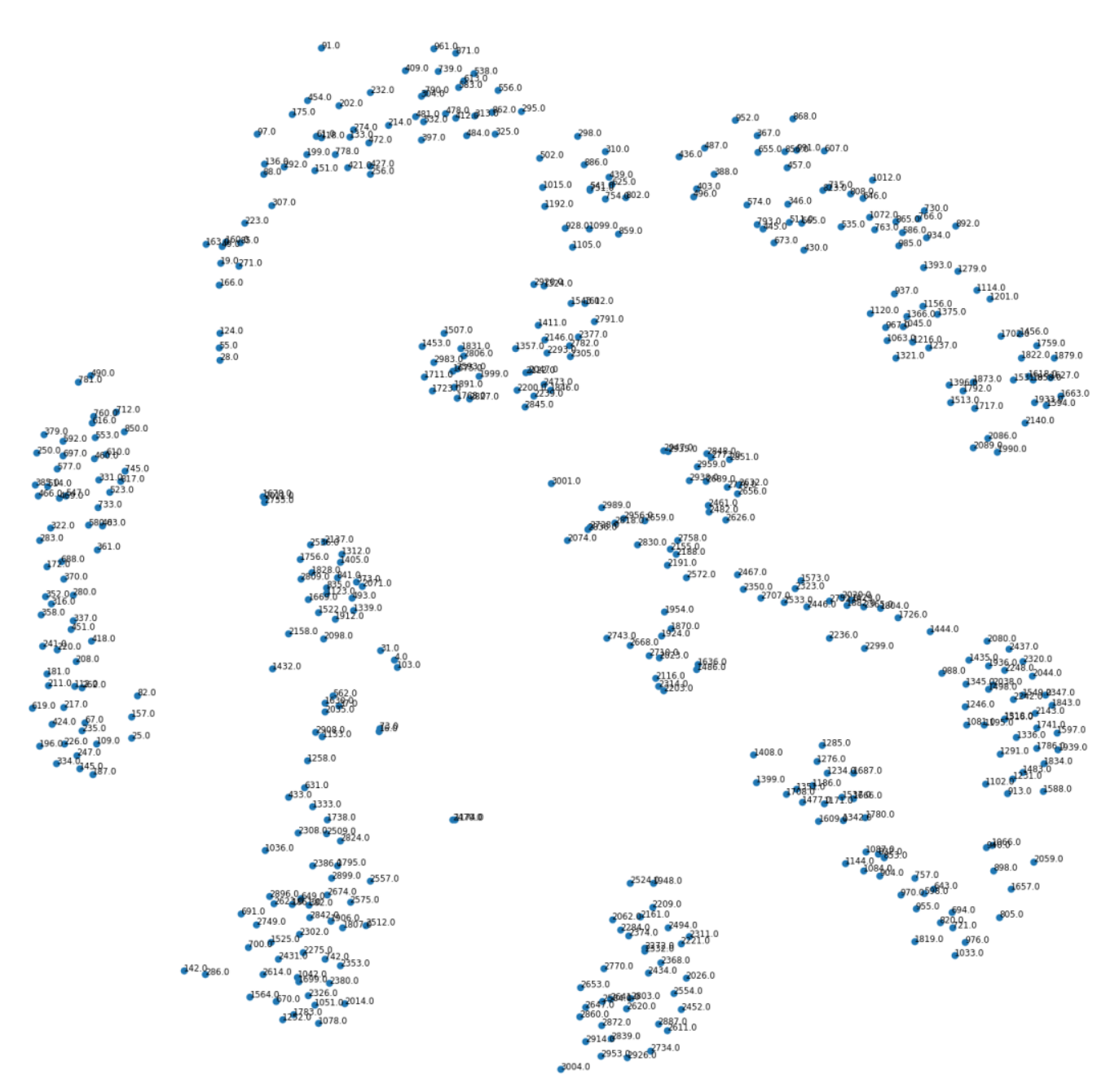}} 
\caption{Visualization of the 460 SRM design candidates optimized at 3 operating points with 13 objectives: (a) visualization by PCA; (b) visualization by Isomap; and (c) visualization by t-SNE.}
\label{motor_16_dim} 
\end{figure*} 

A 3-D model of the switched reluctance machine is shown in Fig. \ref{SRM}(a) with a doubly-salient structure that is simple, low-cost and robust \cite{Shen_J4}. A cross-sectional plot of an $6/4$ SRM is shown in Fig. \ref{SRM}(b), and the performance indices or features of which depend on the geometric parameters indicated in the figure, as well as the excitation current profile. The SRM many-objective (4 or more dimensions) optimization problem in this study is approached by combining the standard NSGA-II optimization algorithm with a proposed analytical model \cite{Shen_J2, Shen_J3} and a multi-objective \cite{Shen_J1, Shen_C3, Shen_C5, Shen_C14} and multi-physics design model \cite{Shen_C10}, where seven prime design variables are specified including the stator bore diameter $D$, the machine stack length $L_{stack}$, the angle span of stator and rotor poles $\theta_s$ and $\theta_r$, the current density $J$, as well as the turn-on and turn-off angles of the excitation current profile $\theta_{on}$ and $\theta_{off}$. A brief flowchart of the adapted analytical optimization and visualization process is presented in Fig. \ref{chart} and more details can be found in \cite{Shen_J1}.
%
 

Specifically, the benchmark SRM is a small-scaled, high speed machine with a $6/4$ typology, rated at 100 W and 10,000 rpm, and thus the efficiency suffers when compared to conventional SRMs because of the drastic increase in iron loss at a high speed, and the intrinsic torque ripple is still large because of its $6/4$ topology. Three operating points of interests are selected to optimize the SRM performance:
\begin{enumerate}
    \item Operating point A: 0.18 N$\cdot$m and 2,000 rpm, with  3 A current excitation;
    \item Operating point B: 0.08 N$\cdot$m and 5,000 rpm, with  2 A current excitation;
    \item Operating point C: 0.02 N$\cdot$m and 10,000 rpm, with 1 A current excitation.
\end{enumerate}

The excitation current is regulated by the hysteresis controllers. During the design and optimization process, the air-gap length and the number of turns in the stator windings are fixed. Other machine design variables, such as the winding AWG size and other geometric parameters that depend on the prime design variables specified earlier, can be calculated on that basis. For ultra-fast calculations, the machine performance indices or features are estimated using an analytical model \cite{Shen_J1, Shen_J2, Shen_J3} with automated scripts, which includes the steel saturation and various commutation effects. Other computational methods, such as the FEA or the simplified FEA, could be also employed, but these require a significantly longer computational time, as it must evaluate hundreds or thousands of design candidates. A many-objective optimization is performed with 20 populations and 50 iterations, which generated 460 design candidates by excluding those that failed to meet the design constraint. In addition, this preservation ratio of 0.46 also showcases the necessity of employing more powerful visualization tools for electric machine designs, since the Pareto front method would become increasingly less discriminative as as the number of objectives increases.

%
%
%

%
%
\subsection{Case Study 1: Visualization SRM Designs Candidates Optimized for a Single Operating Point}
For this case study, the SRM is only optimized for operating point A, and five objectives are selected, namely the average torque, torque density, efficiency, torque ripple, and machine volume. In this scenario, it is already challenging to visualize these five dimensions with either a parallel coordinate plot or a scatter plot of Pareto fronts due to the complexity and a heuristic back-and-forth process to identify and locate the position of each design candidate in these plots. However, standard data clustering methods, such as PCA and Isomap, may still bring valuable insights, since the dimension size (five) is not super large. As can be observed in Fig. \ref{motor_6_dim}(a) and (b), PCA seems to be able to successfully identify six clusters while still left out some outliers, and Isomap also seems to suggest five or six clusters, although some clusters are positioned to be very close to each other. The visualization result of t-SNE is also presented in Fig. \ref{motor_6_dim}(c), where 7 clusters are explicitly presented without any noticeable overlap, and it also cross validated a mediocre visualization performance of PCA and Isomap with a modest objective dimension size. 
\subsection{Case Study 2: Visualization SRM Designs Candidates Optimized at Multi-Operating Points}
In this case study, all three of the operating points are taken into account in the optimization process, and the torque density, average torque, efficiency, and torque ripple for all three points are set as objectives, giving a total of 13 objectives including the machine volume. By observing the visualization results in Fig. \ref{motor_16_dim}, it is obvious that t-SNE has a lot more structure than that offered in PCA and Isomap plots. The 8 clusters are well-separated in this low-dimensional map, and there are fairly larger distances between the clusters when compared to PCA and Isomap, which failed to generate distinguishable clusters.

Starting from the visualization provided by t-SNE, machine designers can obtain some insight on how these design candidates are arranged in the data space, and the centroid of each cluster can be picked to represent other design candidates in the same cluster. Therefore, these picks combined can also well represent all the design candidates in the data space. Starting from here, it is possible to proceed with the next stage of the design, fine-tuning and prototype validation.

\section{Conclusion}
In this paper, the t-SNE algorithm has been successfully applied to visualize the electric machine design candidates optimized at multiple operating points, and these visualizations are significantly better than those produced by other techniques such as PCA and Isomap. By projecting the high-dimensional data onto a low-dimensional map, t-SNE is able to provide more informative insights to machine designers on picking either the initial designs to perform a second round of optimization and fine-tuning, or the final prototype validation.






%

\bibliographystyle{IEEEtran}
\bibliography{IEEEabrv.bib,ref.bib} 
\balance

\balance

\end{document}